\documentclass{isprs} 
\usepackage{placeins}
\usepackage{capt-of}
\usepackage{multicol}
\usepackage{amssymb}
\usepackage{amsmath}
\usepackage{bm}
\usepackage{url}
\usepackage{subfigure}
\usepackage{setspace}
\usepackage{geometry} 
\usepackage{epstopdf}
\usepackage[labelsep=period]{caption}  
\usepackage{graphicx}
\usepackage{algorithm}
\usepackage{algorithmic}
\usepackage{subfigure}
\usepackage{booktabs}
\usepackage{array}
\usepackage{adjustbox}
\usepackage[british]{babel} 
\usepackage[hang]{footmisc}
\usepackage{fancyhdr}
\usepackage{color}

\begin{document}

\title{Investigation of the Challenges of Underwater-Visual-Monocular-SLAM}
\date{}

\author{Michele Grimaldi$^{1,3}$, David Nakath$^{1,2}$, Mengkun She$^{1,2}$, Kevin K\"oser$^{1,2}$}

\address{$^1$Oceanic Machine Vision,
GEOMAR Helmholtz Centre for Ocean Research Kiel,
Wischhofstrasse 1-3, 24148 Kiel, Germany\\
$^2$Marine Data Science, Department of Computer Science, Christian-Albrechts-Universität zu Kiel 24118 Kiel, Germany\\
$^3$Computer Vision and Robotics Research Institute (VICOROB), University
of Girona, 17003 Girona, Spain}

\commission{IV,}{YY} 
\workinggroup{IV/4} 
\icwg{}   

\abstract{In this paper, we present a comprehensive investigation of the challenges of Monocular Visual Simultaneous Localization and Mapping (vSLAM) methods for underwater robots. While significant progress has been made in state estimation methods that utilize visual data in the past decade, most evaluations have been limited to controlled indoor and urban environments, where impressive performance was demonstrated. However, these techniques have not been extensively tested in extremely challenging conditions, such as underwater scenarios where factors such as water and light conditions, robot path, and depth can greatly impact algorithm performance.
Hence, our evaluation is conducted in real-world AUV scenarios as well as laboratory settings which provide precise external reference. A focus is laid on understanding the impact of environmental conditions, such as optical properties of the water and illumination scenarios, on the performance of monocular vSLAM methods. To this end, we first show that all methods perform very well in air and subsequently investigate the degradation of their performance in ever more challenging underwater environments. The final goal of this study is to identify techniques that can improve accuracy and robustness of SLAM methods in such conditions. To achieve this goal, we investigate the potential of image enhancement techniques to improve the quality of input images used by the SLAM methods, specifically in low visibility and extreme lighting scenarios in scattering media. We present a first evaluation on calibration maneuvers and simple image restoration techniques to determine their ability to enable or enhance the performance of monocular SLAM methods in underwater environments.}


\maketitle

\section{Introduction}\label{MANUSCRIPT}
 \begin{figure*}[htb] 
  \centering
  \includegraphics[width=1\textwidth]{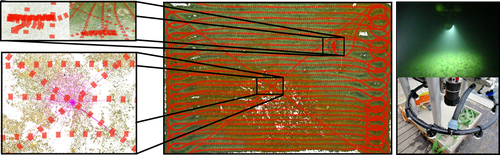}
  \caption{(Middle) Sparse Colmap reconstruction of the real Girona 500 Series AUV \textbf{A}3 surveying mission. After an initialization maneuver, a typical lawn mower patter followed by two cross tracks to support loop closing attempts are executed. The camera trajectory is drawn in red. Left lower magnification: increased number of loop closing opportunities (correspondences drawn in Pink) by executing cross-tracks, while the upper one shows a top- and a side-view of the initialization maneuver carried out to support SLAM algorithms. (Right) AUV in deep underwater mission and deployed camera-light-system.}
  \label{fig:teaser} 
\end{figure*}
\sloppy

Underwater environments present unique challenges for robotic navigation. The low visibility, extreme lighting conditions, and unpredictable nature of underwater terrain make it difficult for robots to accurately perceive their surroundings and navigate effectively. 
Monocular vSLAM (visual Simultaneous Localization and Mapping) methods have emerged as a promising solution for underwater robot navigation, allowing robots to create a map of their surroundings while simultaneously determining their own position within that map \cite{1638022}.

In monocular vSLAM, a single camera is used to capture images of the environment, which are then used to   construct a map of the surrounding area. The camera's position and orientation are estimated in real-time, allowing the robot to determine its own location within the map. However, the accuracy of monocular vSLAM methods can be significantly impacted by environmental conditions, particularly low visibility and extreme lighting scenarios, which are common in underwater environments. To address this challenge, various image enhancement techniques have been proposed to improve the quality of the input images and thereby enhance the performance of monocular vSLAM methods \cite{song2022optical}. These techniques comprise rather heuristic, physically-basedly-based as well as machine learning-based approaches, such as Generative Adversarial Networks (GANs).

In addition to monocular vSLAM methods, sensor fusion algorithms are crucial for accurate and robust navigation of underwater robots. Underwater environments pose unique challenges, such as low visibility and unpredictable terrain, that can significantly impact the accuracy of navigation systems. Therefore, integrating data from multiple sensors, such as Inertial Navigation Systems (INS), sonars, lasers, and cameras, can improve the overall performance of the navigation system.
For example, combining data from an Inertial Navigation System (INS) and a Doppler Velocity Log (DVL) can improve the accuracy of underwater robot navigation by providing velocity measurements that are not affected by currents. Similarly, integrating data from an INS and a sonar can improve underwater localization by providing depth information that can be used to correct INS drift. Sonar sensors, such as multibeam sonars, can also be used to provide high-resolution 3D maps of underwater environments that can be used for localization and mapping \cite{palomer2016multibeam}. Laser-based sensors, such as scanning laser rangefinders, can also be used to provide high-resolution 3D maps of underwater environments that can be used for localization and mapping \cite{Vila20183DUS}. Incorporating data from multiple sensors can be challenging due to the different modalities and measurement noise associated with each particular sensor. However, advancements in sensor fusion algorithms, such as Extended Kalman Filters (EKF) and Unscented Kalman Filters (UKF), have made it possible to effectively combine data from multiple sensors in real-time \cite{WU2019101845}. 

While sensor fusion using multiple modalities has shown great potential in improving underwater robot navigation and mapping, it also comes with additional cost and complexity. Therefore, in this paper, we evaluate the performance of offline and online monocular vSLAM methods for underwater robot navigation in both a real-world and a water tank in a laboratory setting. Our focus is on the impact of environmental conditions, such as water and illumination, on the performance of monocular vSLAM methods.
To this end, we first show the good performance of the chosen SLAM methods in air and then investigate their performance degradation with respect to varying environmental conditions in a scattering medium, i.e., water.
Finally, we investigate the potential of image enhancement techniques to improve accuracy and robustness in challenging underwater environments. 
\vspace{-0.5cm}
\section{Monocular vSLAM}\label{sec:TITLE AND ABSTRACT BLOCK}
Monocular vSLAM methods are commonly classified into three categories: feature-based methods, direct methods, and visual-inertial methods \cite{ZHANG2022100510}. Feature-based methods, such as ORB-SLAM2 \cite{mur2017orb}, rely on detecting and tracking distinctive features in the image frames to estimate camera poses and create a map of the environment. These methods have been shown to achieve accurate results in a variety of scenarios, including indoor and outdoor environments. However, they can be sensitive to changes in illumination, texture, and occlusions, which can cause feature tracking failures and affect their robustness \cite{mur2015orb}. Newer versions of ORB-SLAM2, ORB-SLAM3 \cite{ORBSLAM3_TRO} incorporate semantic information to improve the robustness and accuracy of the system.  Direct methods, such as DSO \cite{engel2018direct} and LSD-SLAM \cite{engel2014lsd} estimate camera motion and 3D structure directly from the intensity values of the image frames, without relying on feature detection and tracking. LSD-SLAM uses semi-dense depth maps to estimate the camera poses and create a map of the environment. This method is known for its ability to handle large-scale environments and low-texture scenes while ORB-SLAM2 uses ORB features to detect and track keypoints in the image frames, and then estimates the camera poses and creates a map of the environment based on the feature matches. BAD-SLAM \cite{schops2019bad} is another direct method that estimates the camera motion and 3D structure directly from the intensity values of the image frames. It aims to improve the accuracy and robustness of SLAM systems by directly leveraging the RGB-D information and performing real-time bundle adjustment.

These methods can provide accurate and dense reconstructions in low-texture environments, but they require significant computational power and are less robust to changes in illumination and scene geometry \cite{engel2017direct}.Visual-inertial methods, such as OKVIS \cite{leutenegger2015keyframe} and VINS-Mono \cite{qin2018vins}, fuse camera and inertial measurements to estimate camera poses and create a map of the environment. These methods can achieve accurate results in highly dynamic and challenging environments, but they require additional sensors and calibration \cite{forster2017manifold}. However, also these methods face challenges when operating in highly dynamic underwater or low-light environments \cite{koser2020challenges}. To address these challenges, researchers have proposed various modifications to existing methods or developed entirely new methods. For example, \cite{s19030687} proposed a new visual odometry method specifically designed to handle the challenging conditions of underwater environments without any previous image enhancement step. 
However, in our study we focus only monocular vSLAM methods that use camera data alone and do not incorporate inertial measurements.
\vspace{-0.3cm}
\section{Image Enhancement / Restoration}
Underwater environments typically exhibit extreme light conditions and poor visibility, which can significantly impair the performance of monocular vSLAM methods. To mitigate this issue, researchers have proposed various techniques for enhancing underwater images, which we here broadly divided into four categories: heuristic, statistical, physically-based, and machine learning methods (see \cite{song2022optical} for a comprehensive survey).

\subsection{Statistical methods}
Statistical methods for underwater image restoration aim to recover the original image from its degraded underwater version. These methods use statistical models to estimate the degradation factors, such as light attenuation, scattering, and noise, and then use these estimates to restore the image. For instance, \cite{Chiang2012} proposed a wavelength compensation and dehazing approach that estimates the light attenuation coefficient and compensates for color distortion caused by the water medium. In \cite{6247661} the authors proposed an underwater image enhancement method that relies only on the degraded version of the image for input and weight measures. They used two inputs to represent color correction and contrast enhancement of the original underwater image/frame, while four weight maps were employed to enhance the visibility of distant objects degraded by medium scattering and absorption.
\cite{7761187} proposed a statistical approach that estimates the parameters of a degradation model and inverts the degradation process to restore the original image. 
\subsection{Heuristic methods }
Heuristic-based methods for underwater image restoration exploit specific underwater environment characteristics. The underwater dark channel prior (UDCP) \cite{drews2013transmission} uses the dark channel prior principle to estimate transmission and restore the image. Mohan's adaptive histogram equalization (AHE) \cite{MOHAN2020941} applies histogram equalization to small image regions for contrast enhancement. Pizer et al.'s contrast-limited adaptive histogram equalization (CLAHE) \cite{pizer1987adaptive} limits contrast enhancement to prevent over-amplification of noise.
In  \cite{köser2021robustly}, the authors present a practical approach to compensating for these lighting effects on flat seafloor regions found in the Abyssal plains. The method is parameter-free
and performs robust statistics-based estimates of additive and multiplicative nuisances without requiring explicit parameters for light, camera, water, or scene. Although heuristic models can produce impressive outcomes in situations that align with their underlying assumptions, their results may lack consistency in other scenarios. Moreover, they do not assure to adhere to physical principles.
\subsection{Machine Learning methods}
Machine learning-based methods have shown promising results in enhancing underwater images by learning from a large dataset of annotated images. One popular machine learning-based method is the Underwater-CNN (UWCNN) proposed in \cite{anwar2018deep}. UWCNN uses a convolutional neural network (CNN) to learn the mapping between the low-quality input underwater image and the high-quality output image. Generative Adversarial Networks (GANs) are a type of deep neural network that consists of a generator and a discriminator. GANs have been used to learn the mapping between degraded and enhanced underwater images. The UWGAN proposed by Li et al. \cite{7995024} aims to improve the visibility of underwater images. It uses an underwater image dataset to train the generator to generate enhanced versions of degraded underwater images. 
Other GAN-based methods for underwater image enhancement include the Conditional GAN (CGAN) \cite{YANG2020115723} and the Multi-Scale GAN \cite{8730425}. Despite benefiting from the expressive capabilities of neural networks, machine learning methods are typically trained under specific conditions. However, underwater environments are often characterized by dynamic conditions and their unpredictability, which can pose challenges to these methods.
\subsection{Physically-based methods}
Physically-based methods for underwater image restoration aim to model the physically-based processes that cause image degradation, such as light attenuation, scattering, and absorption, and then invert these models to restore the image.
In doing so, they are the only methods able to do image restoration as opposed to image enhancement.
These methods typically require knowledge of the physical properties of parts of the scene and can be computationally intensive.
\cite{SHARMA2022101059} proposed a graph-based algorithm for color correction of underwater images. The technique models color relationships via a graph structure and applies root-filtering in graph spectral domains, enhancing high-frequency details for improved visual quality.
Another method is the Sea-thru method proposed by Akkaynak and Treibitz \cite{Akkaynak_2019_CVPR} which estimates the backscatter using the dark pixels and their known range information, and then uses an estimate of the spatially varying illuminant to obtain the range-dependent attenuation coefficient.
The latter method, however, is only valid for the Sun-light case, which exhibits homogeneous illumination.
\cite{boittiaux2023sucre} implemented multiview extensions, to improve the estimates and applied the method to datasets with artificial illumination. However, to make this approach work, the artificial illumination has to be locally homogeneous.
Further approaches, which can be applied to true heterogeneous underwater artificial scenarios, i.e., with artificial illumination, are presented in \cite{https://doi.org/10.1002/rob.21638} and \cite{nakath2021situ}. While physically-based methods have the advantage of being based on well-established principles, they can be limited by the availability and accuracy of the precise parameters needed for the models.
\vspace{-0.3cm}
\section{Datasets}
\begin{figure}[t] 
  \centering
  \includegraphics[width=.95\columnwidth]{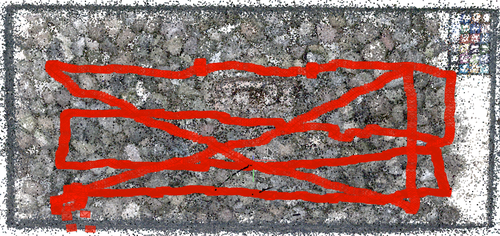} 
  \caption{Top view of a sparse Colmap reconstruction of an example trajectory (in red) in the water tank: in the lower left, we exhibit an initialization maneuver (wiggle over one point), then we execute a lawn mower pattern, and finally cross the tracks, to improve the loop closing impact.} 
  \label{fig:e1sparse} 
\end{figure}
With Girona 500 series AUVs, we collected \textbf{A}-datasets in real waters, as well as \textbf{T}-datasets in a water tank with a precise ground truth estimate. 
In all settings, we carried out dedicated calibration maneuvers, which foster the initialization of SLAM approaches. Furthermore, in the tank and the AUV sets, we carried out classical lawn mower patterns with a stable flying height with subsequent cross-tracks, to support loop-closing approaches. In the tank, we additionally recorded sets, which resemble a more free-flying scanning path. The latter brings about a lot of loop closing opportunities, which will however be impaired by big height variances, which in turn induce big changes in visual appearance in scattering media.

\subsection{Water Test Tank with Ground Truth}

We equipped a $2.2 \times1 \times 0.8$m  water test tank with three 50w Wasler daylight bulbs (5400k) housed in Walimex diffusors to create a homogeneous illumination setting akin to heavy atmospheric scattering.
In addition, we attached two Ulanzi L2 Lite (5500k) co-moving lights, to be able to simulate active underwater light systems to a custom-built externally-tracked underwater camera \cite{winkel2023design}.

\begin{figure}
\centering
\subfigure{\includegraphics[width=0.15\textwidth]
{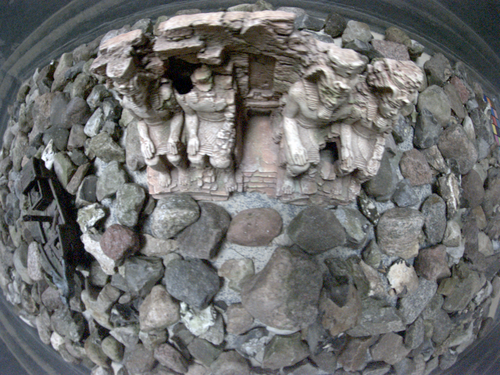}}
\subfigure{\includegraphics[width=0.15\textwidth]
{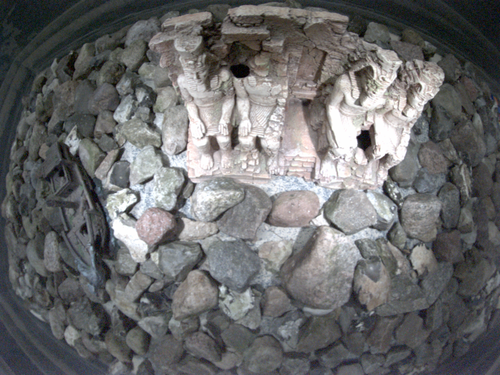}}
\subfigure{\includegraphics[width=0.15\textwidth]{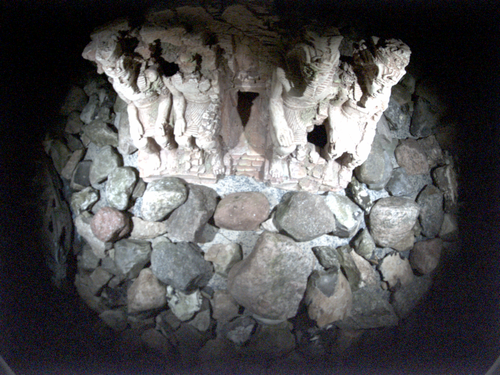}}

\subfigure{\includegraphics[width=0.15\textwidth]
{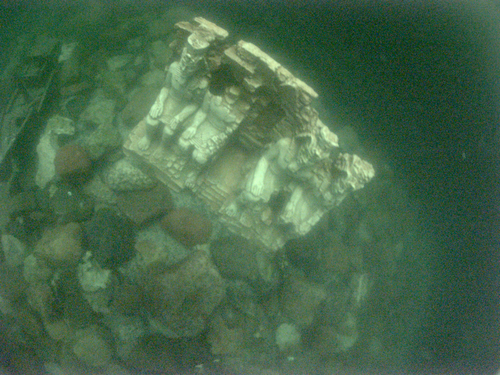}}
\subfigure{\includegraphics[width=0.15\textwidth]
{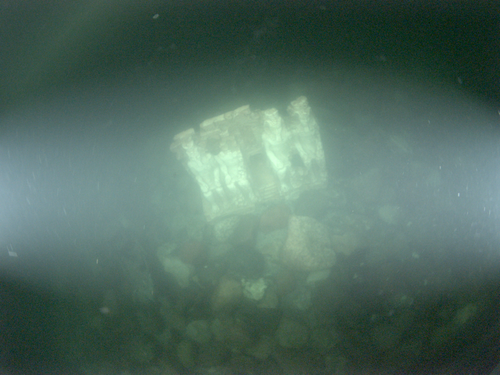}}
\subfigure{\includegraphics[width=0.15\textwidth]{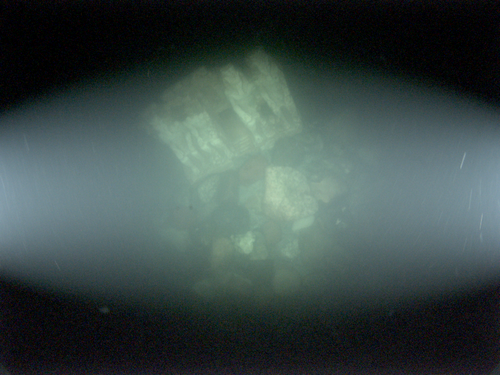}}
\caption{Left to right: Example images of tank sets: with Sunlight, Sun-and artificial light and artificial light. Upper row: in air \textbf{T}1-3, \textbf{T}4-5, and \textbf{T}6-7. Lower row: underwater   \textbf{T}8-9, \textbf{T}10-11, and \textbf{T}12-13. Please note that the images are still distorted, but shown in sRGB-space for better visibility.}
\label{fig:water_tank_uw}
\end{figure}

After building a small-scale test scene, we took several sets, to acquire underwater imagery with external reference as ground truth. 
As this is close to impossible in real waters, we equipped the camera with a stick and attached two Vive controllers, whose pose (position and attitude) in space can be precisely determined in air. 
This information can be used, to obtain a fused estimate of the pose of the underwater camera.
We found the mean accuracy of the system tracking performance to be smaller 3 mm and 0.3 deg for translation and attitude, respectively \cite{winkel2023design}. For the dataset, an in-air fisheye calibration was conducted, with a residual error of 0.22px.
Then, we center the camera within a dome port, to exclude refraction effects from the dataset, stemming from the traversal of interfaces of media with different optical densities \cite{she2019adjustment,she2022refractive}.
Finally, we conducted an underwater fisheye calibration of the camera, with residual reprojection error of 0.55px to capture remaining disturbances, which have not been captured by the preceding steps.
Hence, we will exclusively deal with color distortion effects in those datasets.
Specifically, we took homogeneously illuminated sets (\textbf{T}1-3), sets with mixed illumination (\textbf{T}4/5) and finally two sets with co-moving lights (\textbf{T}6/7) in air (see Fig. \ref{fig:water_tank_uw}). 
Subsequently, we added water and dye for the attenuation as well as Maaloxan as the scattering agent until the working range was clearly distorted by the corresponding effects. This setup enables us to mimic the underwater conditions in Sunlight (\textbf{T}8/9), a mixed (Sun-artificial) light scenario (\textbf{T}10/11), as well as deep sea conditions, where only the artificial light is visible (\textbf{T}12/13); see Fig. \ref{fig:water_tank_uw}).
All former sets of those pairs execute a lawn mover pattern, while all latter sets execute free scanning trajectories with bigger depth variances.

For evaluation, we undistort the images into canonical pinhole space to provide them to the SLAM algorithms.
Their results are then compared to the ground truth provided by the external reference system. 

\subsection{AUV Datasets}

\begin{figure}
\centering
\subfigure{\includegraphics[width=0.15\textwidth]{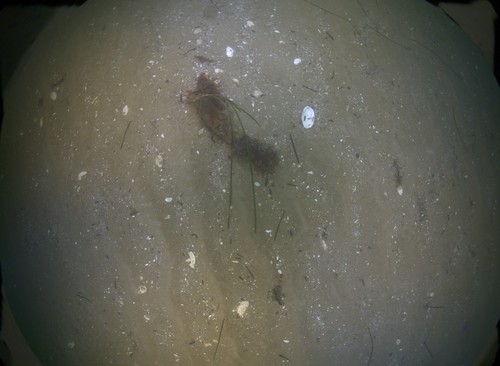}}
\subfigure{\includegraphics[width=0.15\textwidth]{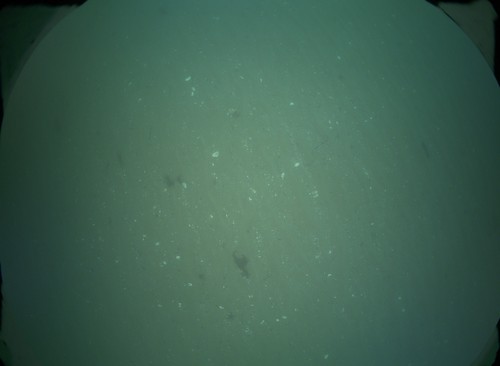}}
\subfigure{\includegraphics[width=0.15\textwidth]{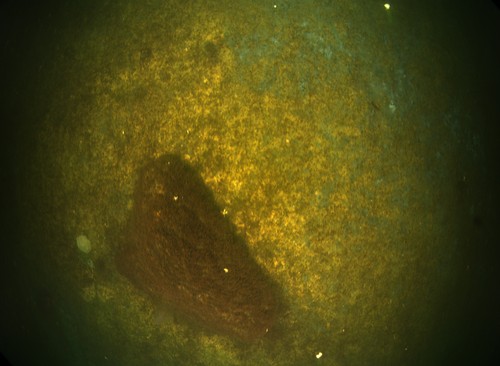}}
\caption{Left to right: Example images from \textbf{A}1, \textbf{A}2, and \textbf{A}3.}
\label{fig:auv_sets}
\end{figure}

We also collected three challenging real datasets with Girona 500 series AUVs in the Baltic Sea. The \textbf{A}1/2 datasets are without initialization maneuvers (see Figs. \ref{fig:auv_sets} a,b), while \textbf{A}3  set features an initialization maneuver tailored to SLAM approaches (see Fig. \ref{fig:auv_sets} c). 

The AUVs have a circular-arranged active lighting system, comprised of 8 LED-compounds cast in rasin (see Fig. \ref{fig:teaser}) \cite{sticklus2017method,song2021deep}.
Specifically, the AUVs are equipped with a dome port camera, which was again calibrated in air with a fisheye model.
Subsequently, it was centered \cite{she2019adjustment,she2022refractive} to avoid refraction effects.
We then overtake the underwater fisheye parameters estimated by Metashapes Photoscan into an adapted Colmap \cite{schoenberger2016sfm} version to establish ground truth with an offline reconstruction method. 
In the latter process, the navigation data is fused into the visual reconstruction using a pose-graph-approach \cite{she2023efficient}.
For evaluation, the images are then undistorted and provided in canonical pinhole space for the SLAM algorithms.
Finally, the results from Colmap's sparse reconstruction serve as the ground truth poses.
\vspace{-0.5cm}
\section{Evaluation}
We evaluated the performance of four SLAM methods, ORB-SLAM2, ORB-SLAM3, LSD-SLAM, and BADSLAM. For each underwater dataset, we applied six different image enhancement methods: the UDCP algorithm, the CLAHE algorithm, UW-GAN algorithm that was trained on three different types of water and the median filter from \cite{köser2021robustly}. We also evaluated each method on the basic, unenhanced images. For BADSLAM and GRADSLAM, we estimated the depth maps using UW-Net \cite{gupta2019unsupervised} and UDepth \cite{yu2022udepth} in underwater scenarios and Monodepth2 \cite{monodepth2} for the in-air sets.

\begin{figure*}[ht]
\centering
\subfigure{\includegraphics[width=0.13\textwidth]{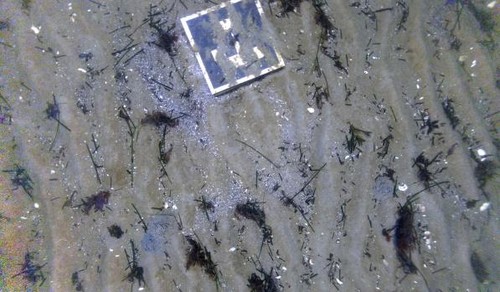}}
\subfigure{\includegraphics[width=0.13\textwidth]{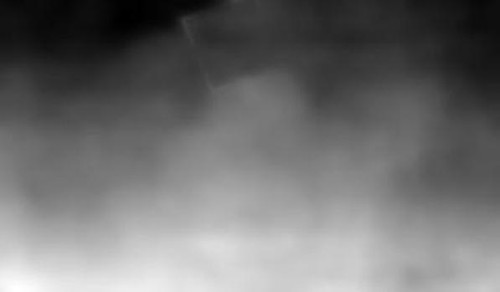}}
\subfigure{\includegraphics[width=0.13\textwidth]{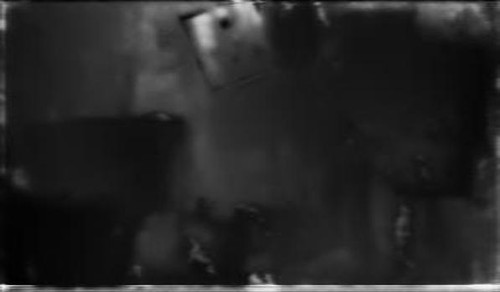}}
\subfigure{\includegraphics[width=0.13\textwidth]{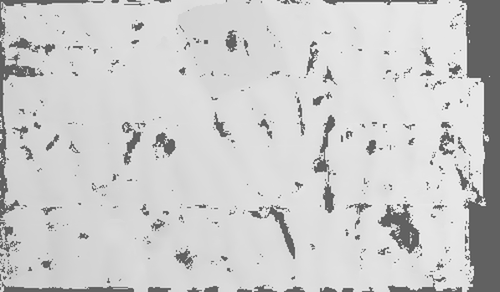}}
\hspace{0.2cm}
\subfigure{\includegraphics[width=0.13\textwidth]{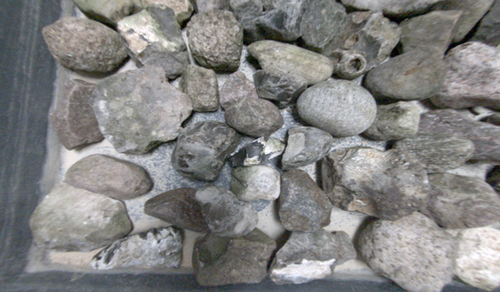}}
\subfigure{\includegraphics[width=0.13\textwidth]{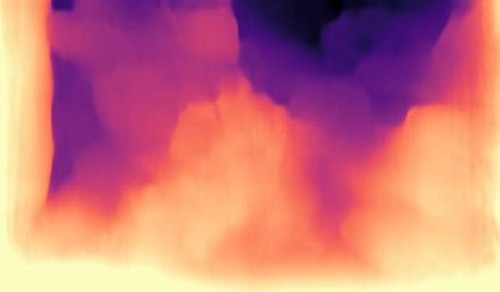}}
\subfigure{\includegraphics[width=0.13\textwidth]{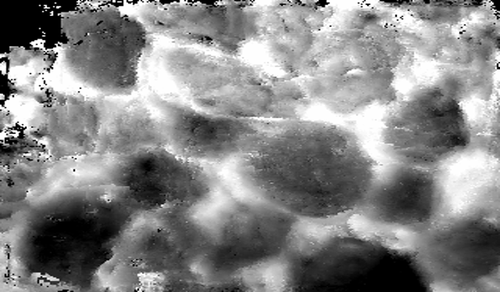}}
\caption{From left to right:  base image from an \textbf{A}-set, UDepth depth map, UW-Net depth map, Colmap depth map; base image from \textbf{T}-set, Monodepth2 depth map, Colmap depth map}\label{fig:dmaps}
\end{figure*}

We categorized failures into three types: not initializing (NOT INIT), initializing but losing track (TR-Lost), and complete failure (FAILED). Not initializing means the method could not start tracking the camera pose. Initializing but losing track indicates that the method began tracking but eventually lost the camera pose without recovery. If the camera poses are lost but the map has enough keyframes, the algorithm is considered successful. Complete failure means the method provided no output (e.g., LSD-SLAM). BADSLAM may fail to start if the estimated depth map is inaccurate. 

We used a solid alignment approach to match up the SLAM estimates and the ground truth. We initially temporally aligned the SLAM estimates and the ground truth interpolating the latter such that for each ground truth pose there is a corresponding estimated pose. Afterwards, we employed the $\mathbb{SIM}3$ \cite{allen2014motion} Umeyama \cite{88573} alignment technique, which considers scale, translation, and rotation, to accurately align the result trajectory with the ground truth in space.
Specifically, we optimize
\begin{equation}
\min_{s, R, t} \sum_{i=1}^{k} \left\| \hat{x}_{i} - (sR x_i + t) \right\|_2^2, s \in  \mathbb{R}; t  \in \mathbb{R}^3; R \in \mathbb{SO}3    
\end{equation}
where $\hat{x_i}$ and $x_i \in \mathbb{R}^3$ are the paired positions. This approach also entails an error-measure in the units of the ground-truth-data, i.e., position in [m] and attitude in [deg]. We used the \cite{grupp2017evo} Python package to perform the alignment.
If the method was able to successfully initialize and track the camera pose, we used the absolute trajectory error (ATE) to compare the ground truth trajectory with the estimated trajectory. The ATE is calculated by finding the difference between the ground truth and estimated camera poses at each frame and then computing the root mean squared error (RMSE) of these differences. This allowed us to compare the accuracy of the SLAM methods under different conditions.
\newline
The ATE can be further broken down into the error for translation and the error for rotation. The translation error measures the difference between the estimated and ground truth position of the camera, while the rotation error measures the difference between the estimated and ground truth orientation of the camera. Both of these errors are calculated using the same approach as for the ATE, by finding the difference between the ground truth and estimated values at each frame and then computing the RMSE of these differences. The ATE errors for translation and rotation are computed as follows:
\begin{equation} \label{eq1}
\begin{split}
\textrm{ATE}_{\textrm{t}} &= \sqrt{\frac{1}{N} \sum_{i=1}^N ||p_i - \hat{p_{i}}||^2} \textrm{~~, and} \\
\textrm{ATE}_{\textrm{r}} &= \sqrt{\frac{1}{N} \sum_{i=1}^{N} 2 \cdot \arccos \left(\min(|q_{i}^{-1}\cdot\hat{q_{i}}|, 1)\right) \cdot \frac{180}{\pi}},
\end{split}
\end{equation}
where $N$ is the number of poses in the trajectory, $p_i, \hat{p}_i \in \mathbb{R}^3$ and $q_i, \hat{q}_i \in \mathbb{SO}3$ are the ground truth and estimated positions and attitudes of the robot at pose $i$.
\vspace{-0.3cm}
\section{Results}

According to the findings presented in Annex A, the experiments conducted on the \textbf{A}1, \textbf{A}2, and \textbf{A}3 datasets revealed that the algorithms encountered challenges and exhibited poor performance due to various factors. Specifically, for the \textbf{A}1 dataset, one of the main issues identified was the lack of sufficient overlap between frames. This insufficient overlap hindered the algorithms' ability to establish robust correspondences and accurately estimate the robot's trajectory.
On the other hand, for the \textbf{A}2 dataset, although there was a good overlap between frames, the presence of unfavorable water and light conditions posed significant difficulties for the algorithms. These conditions, such as poor visibility, light scattering, and limited lighting, adversely affected the algorithms' ability to accurately estimate depth and track the robot's movement. 
Furthermore, as for the case of the \textbf{A}3 dataset, the challenges were further compounded by the presence of low texture areas in addition to the water and lighting conditions. Low texture areas, which lack distinctive visual features, make it challenging for SLAM algorithms to establish reliable correspondences and accurately estimate the robot's trajectory in those regions. In fact, while the initialization trajectory of \textbf{A}3 helped for the initialization of the ORB-SLAMS and the median method performed the best on the \textbf{A}2 dataset, the ORB-SLAMS lost track in the majority of the frames. 

The experiments on the \textbf{T}8-13 dataset, which consisted of a tank with water, have shown that light conditions are critical for successful SLAM performance. The light cones produced from the artificial lights, with and without the sunlight, had a significant impact on the images, resulting in failure of the SLAM methods. This is due to the fact that the presence of water causes light to being refracted, attenuated and scattered. While we controlled for the refraction effects, by centering the camera in the dome, the two latter effects led to changes in image appearance and the degradation of image quality. In addition, the absorption and scattering of light in water varies depending on the wavelength and scene depth, which can affect the accuracy of visual odometry and feature tracking.

Our findings indicated that LSD-SLAM was ineffective for underwater applications, as it failed to function properly on every underwater dataset we tested, consistent with previous research \cite{8968049}. Furthermore, while the method was effective on the sunlight scenarios (\textbf{T}1-\textbf{T}3), it was ineffective on the in-air sets with mixed lights and in-air sets with only artificial light which is also consistent with previous research \cite{pascoe2017nid}. Additionally, our findings highlight the impact of robot maneuvers, both during the trajectory and during the initialization phase, on SLAM accuracy. Specifically, we observed that low dynamic maneuvers tended to result in better accuracy for SLAM. When the robot moved in a more controlled and stable manner, the SLAM algorithm was able to more accurately estimate the robot's position and orientation. Furthermore, the initialization maneuver also had an impact on SLAM accuracy. By carefully designing and executing an appropriate initialization maneuver, we were able to improve the accuracy of the SLAM algorithm. This initialization maneuver provided the algorithm with a more accurate starting point, allowing it to establish a better understanding of the environment and subsequently improve the overall trajectory estimation.

The BADSLAM and GRADSLAM algorithms did not work with either the UDepth or UW-Net depth estimators in underwater scenarios and with Monodepth2 for the in-air sets. To ascertain whether the depth estimators were the root of the issue, we used Colmap for depth estimation and discovered that BADSLAM was capable of at least initializing itself in the \textbf{A}1/2 and \textbf{A}3 datasets and in the \textbf{T} in-air sets it succeeded alongside GRADSLAM. 
In regard to the UW-Net, U-Depth, and Monodepth2 depth estimators, it is important to note that they are primarily designed for forward-looking camera settings and not specifically for top-down views. 
Unsurprisingly, they do not perform well when applied to top-down views, such as those encountered in underwater environments (see Fig. \ref{fig:dmaps}). 

Concerning the in-air sets, our ORB-SLAM results are aligned with results in \cite{otherUnderwaterPaper} in which the authors created multiple datasets using multiple cameras, IMU and a test tank for visual inertial odometry. Finally our experiments also highlighted the importance of using appropriate image enhancement methods for different water conditions. Among the methods tested, CLAHE performed the best overall. The UW-GAN with different water types performed similarly to UDCP, and the second water type of the UW-GAN often performed as well as UDCP, possibly because the water condition used for training  was similar to the water type of the dataset used for evaluation. 
\vspace{-0.3cm}
\section{Conclusion}
In this paper, we conducted an investigation of the challenges of underwater monocular visual SLAM.
To this end, we prepared several AUV-based and controlled lab-datasets. 
All geometric distortions are controlled for in those sets, while they are all taken in extremely low visibility and harsh light conditions. This allows for an in-depth investigation of the impact of radiometric distortions in the underwater setting. The ground truth in the lab sets is established with a custom-build external reference system, while the AUV sets -- lacking external reference -- are offline-reconstructed with a modified version of Colmap.
First, we showed that all selected SLAM algorithms successfully run on in air tank-dataets with good performances.
Then, we evaluated several combinations of SLAM systems and preprocessing methods on all datasets. 
We found that no SLAM system is able to complete the real AUV-Datasets.
Also from the tank datasets, only the homogeneously illuminated Sun-settings could be completed.
Here, we found that the pre-processing approaches showed some initial improvements of the SLAM performance in the visually adversarial underwater environments.
In addition, we can also report mild improvements, when special initialization-maneuvers are carried out.
The generalization of the pre-processing methods is a direction worth to further investigate, as they seem to be heavily tuned to certain assumptions / scenarios.
Furthermore, providing depth-information dependent SLAM systems with corresponding top-down-view estimates also seems to be an interesting route. 
However we had to resort to depth maps established in an offline fashion, as the deep learning based estimators were tuned to different use cases.
Hence, in the future, we will strive to preprocess underwater imagery to mitigate radiometric distortions and at the same time improve on underwater monocular depth estimation in order to leverage the already existing big potential of in-air SLAM approaches.
\vspace{-0.3cm}
\section{Acknowledgment}

This publication has been funded by the German Research Foundation (Deutsche Forschungsgemeinschaft, DFG) Projektnummer 396311425, through the Emmy Noether Programme. Furthermore, the authors would like to thank Tim Benedikt von See, Jens Greinert, and GEOMAR's AUV team for collecting the AUV datasets within the scope of the BMBF funded CONMAR project (grant no. 03F0912A) and ships proposal MeCoMM (GPF22-1/015)  and Birger Winkel for support while recording the VIVE-tracked tank-data.
\vspace{-0.3cm}
\begin{spacing}{1.17}
    \normalsize
    \bibliography{ISPRSguidelines_authors} 
\end{spacing}

 \appendix
 \newgeometry{
 a4paper,
 total={170mm,257mm},
 left=5mm,
 right=5mm,
 top=5mm,
 bottom=1mm
 }
\twocolumn
 \begin{center}
  \section{Annex Algorithm Performances}
\end{center}
\begin{table}[H]
\centering
\huge
\begin{adjustbox}{width=1\columnwidth}
\begin{tabular}{@{}cccccccc@{}}
\toprule
\multicolumn{8}{c}{\textbf{\textbf{A}1, Real mission w/ Girona 500 AUV}} \\
\midrule
& Base & CLAHE & UDCP & UW-GAN Water 1 & UW-GAN Water 2 & UW-GAN Water 3 & Median \\
\midrule
ORB-SLAM2 & TR-Lost & TR-Lost & TR-Lost & TR-Lost & TR-Lost & TR-Lost & TR-Lost \\
ORB-SLAM3 & TR-Lost & TR-Lost & TR-Lost & TR-Lost & TR-Lost & TR-Lost & TR-Lost \\
BADSLAM (UDepth) & FAILED & FAILED & FAILED & FAILED & FAILED & FAILED & FAILED \\
BADSLAM (UW-Net) & FAILED & FAILED & FAILED & FAILED & FAILED & FAILED & FAILED \\
BADSLAM (Colmap) & TR-Lost & TR-Lost & TR-Lost & TR-Lost & TR-Lost & TR-Lost & TR-Lost \\
LSD-SLAM & FAILED & FAILED & FAILED & FAILED & FAILED & FAILED & FAILED \\
GRADSLAM(UDepth) & FAILED & NA & FAILED & FAILED & FAILED & FAILED & FAILED \\
GRADSLAM(UW-Net) & FAILED & NA & FAILED & FAILED & FAILED & FAILED & FAILED \\
GRADSLAM(Colmap) & FAILED & NA & FAILED & FAILED & FAILED & FAILED & FAILED \\
\bottomrule
\end{tabular}
\end{adjustbox}
\label{tab:sub1}
\end{table}
\vspace{-0.5em}
\begin{table}[H]
\centering
\huge
\begin{adjustbox}{width=1\columnwidth}
\begin{tabular}{@{}cccccccc@{}}
\toprule
\multicolumn{8}{c}{\textbf{\textbf{A}2, Real mission w/ Girona 500 AUV}} \\
\midrule
 & Base & CLAHE & UDCP & UW-GAN Water 1 & UW-GAN Water 2 & UW-GAN Water 3 & Median\\
      \midrule
      ORB-SLAM2 & NOT INIT & TR-Lost & TR-Lost & NOT INIT & NOT INIT & NOT INIT  & TR-Lost\\
      ORB-SLAM3  & NOT INIT  & TR-Lost & NOT INIT & NOT INIT & NOT INIT & NOT INIT & TR-Lost\\
      BADSLAM (UDepth) & FAILED & FAILED & FAILED & FAILED & FAILED & FAILED & FAILED \\
      BADSLAM (UW-Net) & FAILED & FAILED & FAILED & FAILED & FAILED & FAILED & FAILED\\
      BADSLAM (Colmap) & NOT INIT & TR-Lost & TR-Lost & NOT INIT & NOT INIT & NOT INIT & TR-Lost\\
      LSD-SLAM  & FAILED & FAILED & FAILED & FAILED & FAILED & FAILED & FAILED \\
      GRADSLAM(UDepth)  & FAILED & NA & FAILED & FAILED & FAILED & FAILED & FAILED\\
      GRADSLAM(UW-Net)  & FAILED & NA & FAILED & FAILED & FAILED & FAILED & FAILED\\
      GRADSLAM(Colmap)  & FAILED & NA & FAILED & FAILED & FAILED & FAILED & FAILED\\
      \bottomrule
    \end{tabular}
    \end{adjustbox}
\label{tab:sub2}
\end{table}
\vspace{-0.5em}
\begin{table}[H]
\centering
\huge
\begin{adjustbox}{width=1\columnwidth}
\begin{tabular}{@{}cccccccc@{}}
\toprule
\multicolumn{8}{c}{\textbf{\textbf{A}3, Real mission w/ Girona 500 AUV}} \\
\midrule
      & Base & CLAHE & UDCP & UW-GAN Water 1 & UW-GAN Water 2 & UW-GAN Water 3 & Median \\
      \midrule
      ORB-SLAM2 & TR-Lost & TR-Lost & TR-Lost & TR-Lost & TR-Lost & TR-Lost & TR-Lost\\
      ORB-SLAM3 & TR-Lost & TR-Lost & TR-Lost & TR-Lost & TR-Lost & TR-Lost & TR-Lost\\
      BADSLAM (UDepth) & FAILED & FAILED & FAILED & FAILED & FAILED & FAILED & FAILED\\
      BADSLAM (UW-Net) & FAILED & FAILED & FAILED & FAILED & FAILED & FAILED & FAILED\\
      BADSLAM (Colmap) & TR-Lost & TR-Lost & TR-Lost & TR-Lost & TR-Lost & TR-Lost & TR-Lost\\
      LSD-SLAM  & FAILED & FAILED & FAILED & FAILED & FAILED & FAILED & FAILED\\
      GRADSLAM(UDepth) & FAILED & NA & FAILED & FAILED & FAILED & FAILED & FAILED\\
      GRADSLAM(UW-Net) & FAILED & NA & FAILED & FAILED & FAILED & FAILED & FAILED\\
      GRADSLAM(Colmap) & FAILED & NA & FAILED & FAILED & FAILED & FAILED & FAILED\\
      \bottomrule
    \end{tabular}
    \end{adjustbox}
\label{tab:sub3}
\end{table}
\vspace{-0.5em}
\begin{table}[H]
\centering
\huge
\begin{adjustbox}{width=1\columnwidth}
\begin{tabular}{@{}cccccccc@{}}
\toprule
\multicolumn{8}{c}{\textbf{\textbf{T}8, Water Tank, homogenous illumination, lawn mower}} \\
\midrule
      & Base & CLAHE & UDCP & UW-GAN Water 1 & UW-GAN Water 2 & UW-GAN Water 3 & Median \\
      \midrule
      ORB-SLAM2 & $3.5^\circ| 1.42$ & \boldsymbol{$2.2^\circ}|1.28$ & $3.12^\circ| 1.41$  & $4.0^\circ | 1.52$ & $4.1^\circ|1.53$ & $3.9^\circ |1.61$ & NOT INIT \\
      ORB-SLAM3 & $4.5^\circ| 1.61$ & $3^\circ|\boldsymbol{1.19$} & $2.6^\circ| 1.51$  & $3.3^\circ | 1.54$ & $3.8^\circ|1.50$ & $3.3^\circ |1.59$ & NOT INIT\\
      BADSLAM (UDepth) & FAILED & FAILED & FAILED & FAILED & FAILED & FAILED & FAILED \\
      BADSLAM (UW-Net) & FAILED & FAILED & FAILED & FAILED & FAILED & FAILED & FAILED\\
      LSD-SLAM  & FAILED & FAILED & FAILED & FAILED & FAILED & FAILED & FAILED\\
      GRADSLAM(UDepth) & FAILED & NA & FAILED & FAILED & FAILED & FAILED & FAILED\\
      GRADSLAM(UW-Net) & FAILED & NA & FAILED & FAILED & FAILED & FAILED & FAILED\\
      \bottomrule
    \end{tabular}
    \end{adjustbox}
\label{tab:sub4}
\end{table}
\vspace{-0.5em}
\begin{table}[H]
\centering
\huge
\begin{adjustbox}{width=1\columnwidth}
\begin{tabular}{@{}cccccccc@{}}
\toprule
\multicolumn{8}{c}{\textbf{\textbf{T}9, Water Tank, homogenous illumination, scanning trajectory}} \\
\midrule
      & Base & CLAHE & UDCP & UW-GAN Water 1 & UW-GAN Water 2 & UW-GAN Water 3 & Median \\
      \midrule
      ORB-SLAM2 & $2.7^\circ| 1.34$ & $2.2^\circ|1.26$ & $2.2^\circ| 1.52$  & $2.7^\circ | 1.47$ & $2.8^\circ|1.46$ & $2.9^\circ |1.61$ & NOT INIT\\
      ORB-SLAM3 & $2.9^\circ| 1.53$ & \boldsymbol{$1.8^\circ|1.24$} & $2.6^\circ| 1.51$  & $2.62^\circ | 1.49$ & $3.32^\circ|1.59$ & $2.9^\circ |1.47$ & NOT INIT \\
      BADSLAM (UDepth) & FAILED & FAILED & FAILED & FAILED & FAILED & FAILED & FAILED \\
      BADSLAM (UW-Net) & FAILED & FAILED & FAILED & FAILED & FAILED & FAILED & FAILED\\
      LSD-SLAM  & FAILED & FAILED & FAILED & FAILED & FAILED & FAILED & FAILED\\
      GRADSLAM(UDepth) & FAILED & NA & FAILED & FAILED & FAILED & FAILED & FAILED\\
      GRADSLAM(UW-Net) & FAILED & NA & FAILED & FAILED & FAILED & FAILED & FAILED\\
      \bottomrule
    \end{tabular}
    \end{adjustbox}
  \label{tab:sub5}
\end{table}
\vspace{-0.5em}
\begin{table}[H]
\centering
\huge
\begin{adjustbox}{width=1\columnwidth}
\begin{tabular}{@{}cccccccc@{}}
\toprule
\multicolumn{8}{c}{\textbf{\textbf{T}10, Water Tank, mixed illumination, lawn mower}} \\
\midrule
      & Base & CLAHE & UDCP & UW-GAN Water 1 & UW-GAN Water 2 & UW-GAN Water 3 & Median \\
      \midrule
      ORB-SLAM2 & NOT INIT & TR-Lost & NOT INIT  & NOT INIT & NOT INIT & NOT INIT & NOT INIT\\
      ORB-SLAM3  & NOT INIT & TR-Lost & NOT INIT & NOT INIT & NOT INIT & NOT INIT & NOT INIT\\
      BADSLAM (UDepth) & FAILED & FAILED & FAILED & FAILED & FAILED & FAILED & NOT INIT\\
      BADSLAM (UW-Net) & FAILED & FAILED & FAILED & FAILED & FAILED & FAILED & FAILED\\
      LSD-SLAM  & FAILED & FAILED & FAILED & FAILED & FAILED & FAILED & FAILED\\
      GRADSLAM(UDepth) & FAILED & NA & FAILED & FAILED & FAILED & FAILED & FAILED\\
      GRADSLAM(UW-Net) & FAILED & NA & FAILED & FAILED & FAILED & FAILED & FAILED\\
      \bottomrule
    \end{tabular}
    \end{adjustbox}
\label{tab:sub6}
\end{table}
\vspace{-0.5em}
\begin{table}[H]
\centering
\huge
\begin{adjustbox}{width=1\columnwidth}
\begin{tabular}{@{}cccccccc@{}}
\toprule
\multicolumn{8}{c}{\textbf{\textbf{T}11, Water Tank, mixed illumination, scanning trajectory}} \\
\midrule
 & Base & CLAHE & UDCP & UW-GAN Water 1 & UW-GAN Water 2 & UW-GAN Water 3 & Median\\
      \midrule
      ORB-SLAM2 & NOT INIT & TR-Lost & NOT INIT  & NOT INIT & NOT INIT & NOT INIT & NOT INIT\\
      ORB-SLAM3  & NOT INIT & TR-Lost & NOT INIT & NOT INIT & NOT INIT & NOT INIT & NOT INIT\\
      BADSLAM (UDepth) & FAILED & FAILED & FAILED & FAILED & FAILED & FAILED & NOT INIT\\
      BADSLAM (UW-Net) & FAILED & FAILED & FAILED & FAILED & FAILED & FAILED & FAILED\\
      LSD-SLAM  & FAILED & FAILED & FAILED & FAILED & FAILED & FAILED & FAILED \\
      GRADSLAM(UDepth) & FAILED & NA & FAILED & FAILED & FAILED & FAILED & FAILED\\
      GRADSLAM(UW-Net) & FAILED & NA & FAILED & FAILED & FAILED & FAILED & FAILED\\
      \bottomrule
    \end{tabular}
    \end{adjustbox}
\label{tab:sub7}
\end{table}
\vspace{-0.5em}
\begin{table}[H]
\centering
\huge
\begin{adjustbox}{width=1\columnwidth}
\begin{tabular}{@{}cccccccc@{}}
\toprule
\multicolumn{8}{c}{\textbf{\textbf{T}12, Water Tank, artificial illumination, lawn mower}} \\
\midrule
 & Base & CLAHE & UDCP & UW-GAN Water 1 & UW-GAN Water 2 & UW-GAN Water 3 & Median\\
      \midrule
      ORB-SLAM2 & NOT INIT & TR-Lost & NOT INIT  & NOT INIT & NOT INIT & NOT INIT & NOT INIT\\
      ORB-SLAM3  & NOT INIT & TR-Lost & NOT INIT & NOT INIT & NOT INIT & NOT INIT & NOT INIT\\
      BADSLAM (UDepth) & FAILED & FAILED & FAILED & FAILED & FAILED & FAILED & FAILED \\
      BADSLAM (UW-Net) & FAILED & FAILED & FAILED & FAILED & FAILED & FAILED & FAILED\\
      LSD-SLAM  & FAILED & FAILED & FAILED & FAILED & FAILED & FAILED & FAILED\\
      GRADSLAM(UDepth) & FAILED & NA & FAILED & FAILED & FAILED & FAILED & FAILED\\
      GRADSLAM(UW-Net) & FAILED & NA & FAILED & FAILED & FAILED & FAILED & FAILED\\
      \bottomrule
    \end{tabular}
    \end{adjustbox}
\label{tab:sub8}
\end{table}
\vspace{-0.5em}
\begin{table}[H]
\centering
\huge
\begin{adjustbox}{width=1\columnwidth}
\begin{tabular}{@{}cccccccc@{}}
\toprule
\multicolumn{8}{c}{\textbf{\textbf{T}13, Water Tank, artificial illumination, scanning trajectory}}\\
\midrule
  & Base & CLAHE & UDCP & UW-GAN Water 1 & UW-GAN Water 2 & UW-GAN Water 3 & Median\\
      \midrule
      ORB-SLAM2 & NOT INIT & TR-Lost & NOT INIT  & NOT INIT & NOT INIT & NOT INIT & NOT INIT\\
      ORB-SLAM3  & NOT INIT & TR-Lost & NOT INIT & NOT INIT & NOT INIT & NOT INIT & NOT INIT\\
      BADSLAM (UDepth) & FAILED & FAILED & FAILED & FAILED & FAILED & FAILED & FAILED\\
      BADSLAM (UW-Net) & FAILED & FAILED & FAILED & FAILED & FAILED & FAILED & FAILED\\
      LSD-SLAM  & FAILED & FAILED & FAILED & FAILED & FAILED & FAILED & FAILED\\
      GRADSLAM(UDepth) & FAILED & NA & FAILED & FAILED & FAILED & FAILED & FAILED\\
      GRADSLAM(UW-Net) & FAILED & NA & FAILED & FAILED & FAILED & FAILED & FAILED\\
      \bottomrule
    \end{tabular}
    \end{adjustbox}
\label{tab:sub9}
\end{table}
\vspace{-0.5em}
\begin{table}[H]
\centering
\huge
\begin{adjustbox}{width=1\columnwidth}
\begin{tabular}{@{}cccccccc@{}}
\toprule
\multicolumn{8}{c}{\textbf{T}1-7, Tank in-air} \\
\midrule
      & \textbf{T}1 & \textbf{T}2 & \textbf{T}3 & \textbf{T}4 & \textbf{T}5 & \textbf{T}6 & \textbf{T}7 \\
      \midrule
      ORB-SLAM2 & $1.68^\circ | 0.09$ & $1.31^\circ |0.08$  & $2.12^\circ|0.12$ & $1.68^\circ |0.10$  &  $2.47^\circ|0.15$ & $2.23^\circ| 0.21$ & $2.54^\circ |0.16$\\
      ORB-SLAM3  & $1.60^\circ| 0.09$ & $1.23^\circ| \boldsymbol{0.07}$ & $2.94^\circ |0.17$ & $1.92^\circ| 0.12$ & $2.27^\circ | 0.14$ & $2.22^\circ| 0.30$ & $2.23 ^\circ | 0.29$ \\
      LSD-SLAM  & $1.89^\circ| 0.10$ & $1.68^\circ| 0.09$ & $1.98^\circ |0.20$ & $2.20^\circ| 0.24$ & $1.87^\circ | 0.20$ & FAILED & FAILED \\
      BADSLAM (Monodepth2) & FAILED & FAILED & FAILED & FAILED & FAILED & FAILED & FAILED \\
      BADSLAM (Colmap) & $\boldsymbol{1.38^\circ} | 0.08$  &$1.47^\circ | 0.11$ & $1.87^\circ | 0.12$& $1.94^\circ | 0.20$& $2.09 ^\circ | 0.25$&  $2.07^\circ | 0.19$& $2.17 ^\circ | 0.23$\\
      GRADSLAM (Monodepth2) & FAILED & FAILED & FAILED & FAILED & FAILED & FAILED & FAILED \\
      GRADSLAM (Colmap) & $1.45^\circ | 0.12$ & $1.42^\circ | 0.13$  & $2.10^\circ | 0.15$ &$2.22^\circ | 0.17$& $2.3 ^\circ | 0.27$& $2.28^\circ | 0.24$& $2.41 ^\circ | 0.28$\\
      \bottomrule
    \end{tabular}
    \end{adjustbox}
\label{tab:sub10}
\end{table}
\vspace{-0.5cm}
\section{Annex Original and Preprocessed Images}
\begin{figure}[ht]
\centering
\small
\hspace{0.3cm}Original \hspace{0.8cm} CLAHE \hspace{1cm}UDCP \hspace{0.7cm} UWGAN W1 \hspace{0.5cm}
\subfigure{\includegraphics[width=0.09\textwidth]{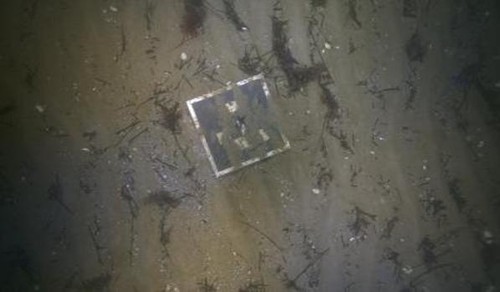}}
\hspace{0.1cm}
\subfigure{\includegraphics[width=0.09\textwidth]{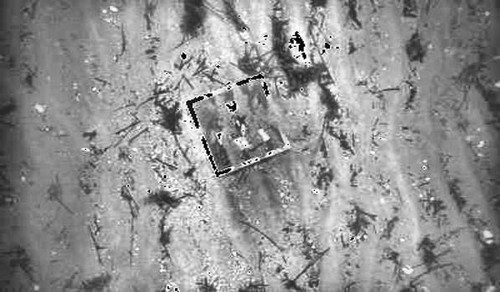}}
\hspace{0.1cm}
\subfigure{\includegraphics[width=0.09\textwidth]{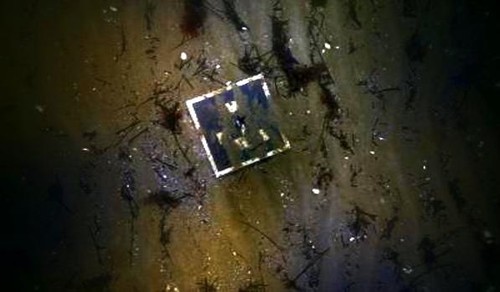}}
\hspace{0.1cm}
\subfigure{\includegraphics[width=0.09\textwidth]{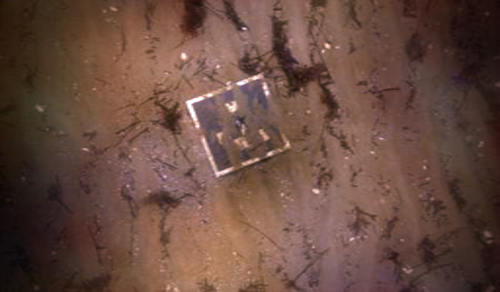}} \hspace{5cm}
\begin{picture}(5,5)
\put(5,5){}
\end{picture}
\hspace{0.3cm} UWGAN W2 \hspace{0.2cm} UWGAN W3 \hspace{0.5cm} Median
\newline
\subfigure{\includegraphics[width=0.09\textwidth]{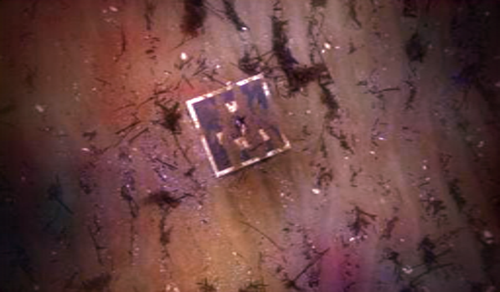}}
\hspace{0.1cm}
\subfigure{\includegraphics[width=0.09\textwidth]{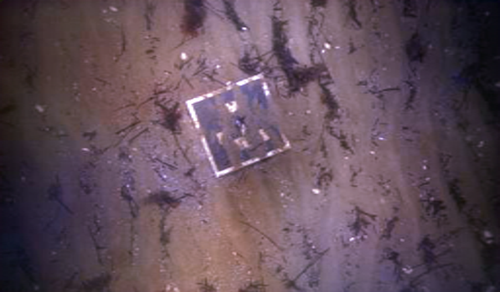}}
\hspace{0.1cm}
\subfigure{\includegraphics[width=0.09\textwidth]{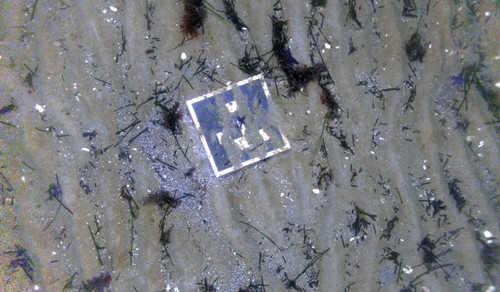}}
\vspace{-0.2cm}
\caption{\textbf{A}1}
\label{fig:six_images_1}
\end{figure}
\begin{figure}[H]
\centering
\hspace{0.3cm}Original \hspace{0.8cm} CLAHE \hspace{1cm}UDCP \hspace{0.7cm} UWGAN W1 \hspace{0.5cm}
\subfigure{\includegraphics[width=0.09\textwidth]{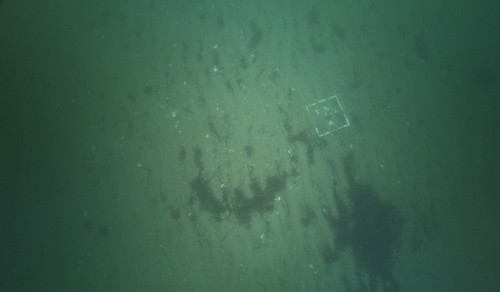}}
\hspace{0.1cm}
\subfigure{\includegraphics[width=0.09\textwidth]{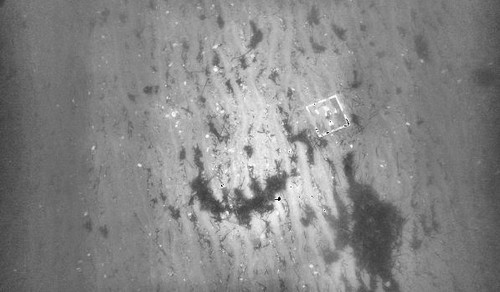}}
\hspace{0.1cm}
\subfigure{\includegraphics[width=0.09\textwidth]{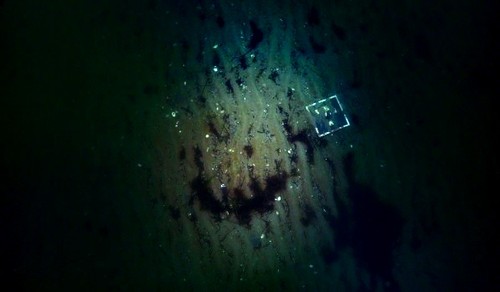}}
\hspace{0.1cm}
\subfigure{\includegraphics[width=0.09\textwidth]{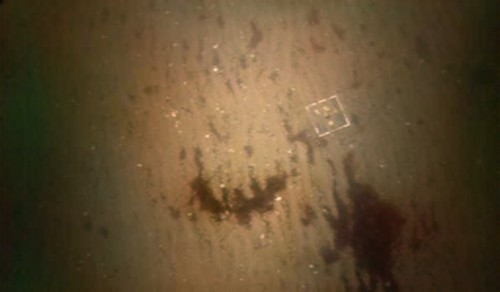}}
\hspace{5cm}
\begin{picture}(5,5)
\put(5,5){}
\end{picture}
\hspace{0.3cm} UWGAN W2 \hspace{0.2cm} UWGAN W3 \hspace{0.5cm} Median
\newline
\subfigure{\includegraphics[width=0.09\textwidth]
{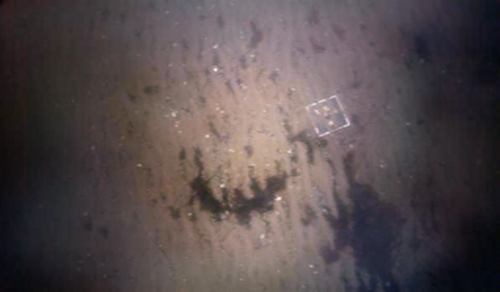}}
\hspace{0.1cm}
\subfigure{\includegraphics[width=0.09\textwidth]{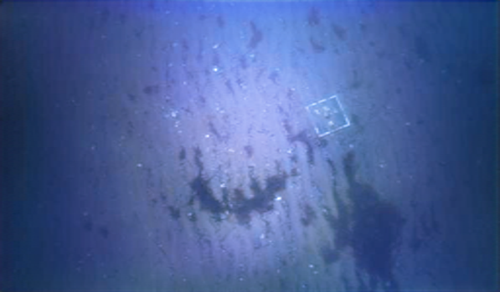}}
\hspace{0.1cm}
\subfigure{\includegraphics[width=0.09\textwidth]{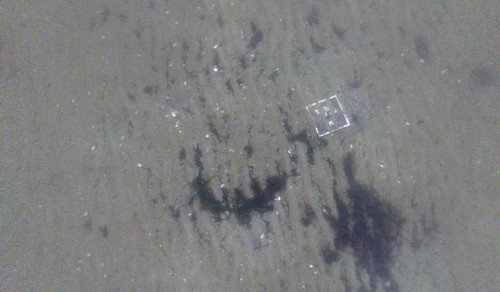}}
\vspace{-0.2cm}
\caption{\textbf{A}2}
\label{fig:six_images_2}
\end{figure}
\vspace{-0.3cm}
\begin{figure}[ht]
\centering
\hspace{0.3cm}Original \hspace{0.8cm} CLAHE \hspace{1cm}UDCP \hspace{0.7cm} UWGAN W1 \hspace{0.5cm}
\subfigure{\includegraphics[width=0.09\textwidth]{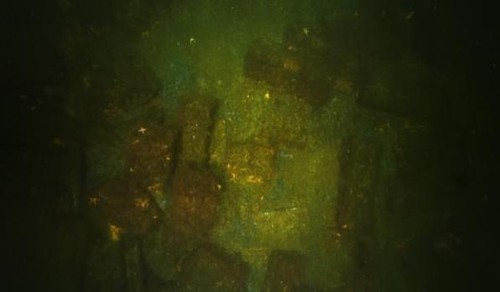}}
\hspace{0.1cm}
\subfigure{\includegraphics[width=0.09\textwidth]{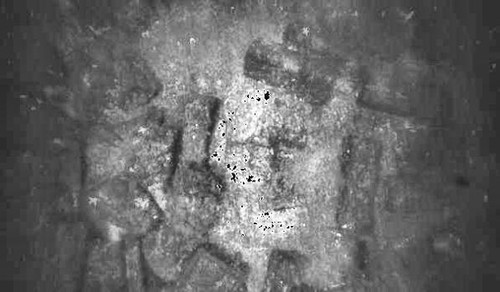}}
\hspace{0.1cm}
\subfigure{\includegraphics[width=0.09\textwidth]{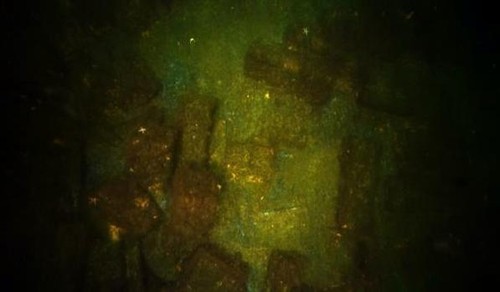}}
\hspace{0.1cm}
\subfigure{\includegraphics[width=0.09\textwidth]{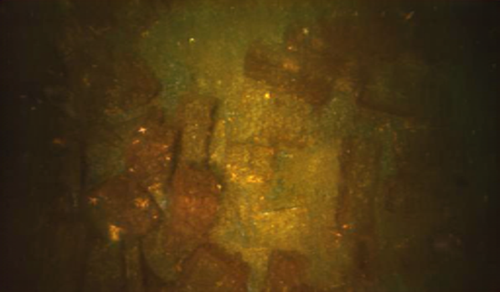}}
\hspace{5cm}
\begin{picture}(5,5)
\put(5,5){}
\end{picture}
\hspace{0.3cm} UWGAN W2 \hspace{0.2cm} UWGAN W3 \hspace{0.5cm} Median
\newline
\subfigure{\includegraphics[width=0.09\textwidth]{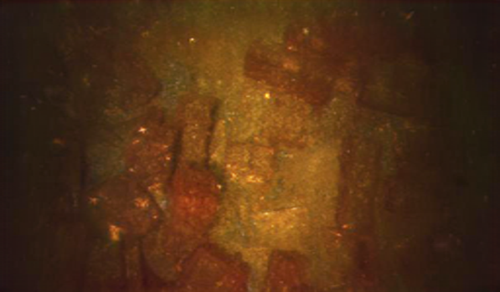}}
\hspace{0.1cm}
\subfigure{\includegraphics[width=0.09\textwidth]{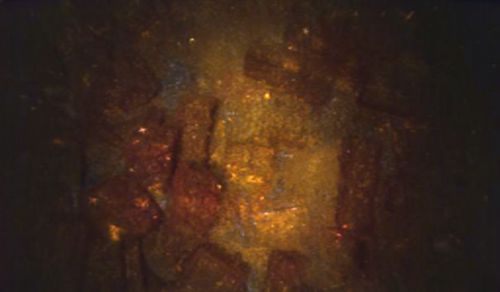}}
\hspace{0.1cm}
\subfigure{\includegraphics[width=0.09\textwidth]{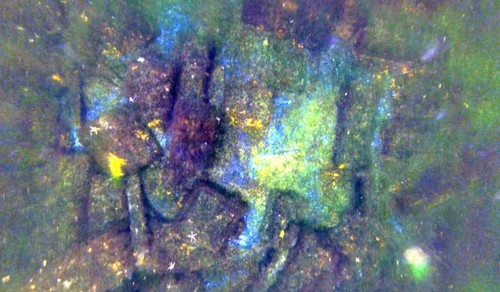}}
\vspace{-0.2cm}
\caption{\textbf{A}3}
\label{fig:six_images_3}
\end{figure}
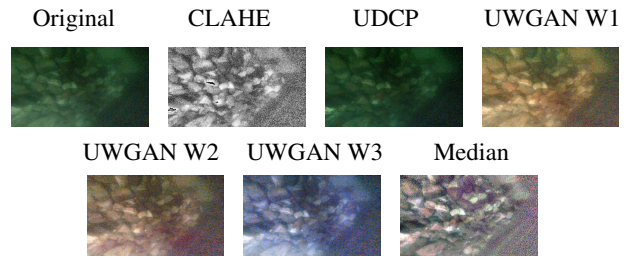
\begin{figure}[H]
\centering
\hspace{0.3cm}Original \hspace{0.8cm} CLAHE \hspace{1cm}UDCP \hspace{0.7cm} UWGAN W1 \hspace{0.5cm}
\subfigure{\includegraphics[width=0.09\textwidth]{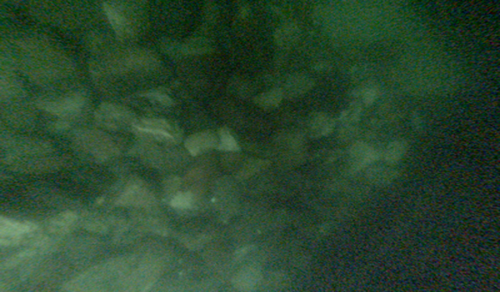}}
\hspace{0.1cm}
\subfigure{\includegraphics[width=0.09\textwidth]{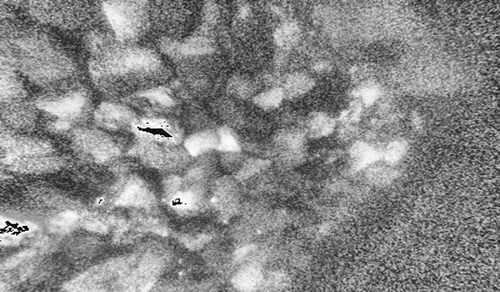}}
\hspace{0.1cm}
\subfigure{\includegraphics[width=0.09\textwidth]{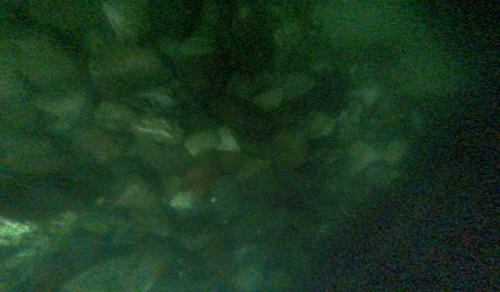}}
\hspace{0.1cm}
\subfigure{\includegraphics[width=0.09\textwidth]{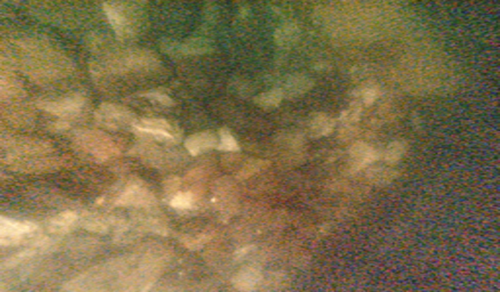}}
\hspace{5cm}
\begin{picture}(5,5)
\put(5,5){}
\end{picture}
\hspace{0.3cm} UWGAN W2 \hspace{0.2cm} UWGAN W3 \hspace{0.5cm} Median
\newline
\subfigure{\includegraphics[width=0.09\textwidth]{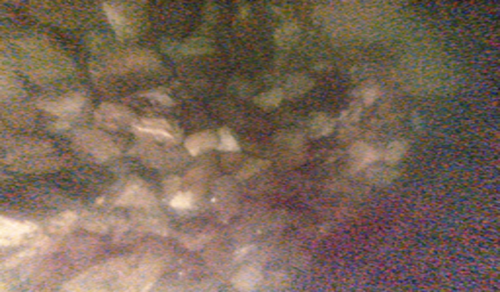}}
\hspace{0.1cm}
\subfigure{\includegraphics[width=0.09\textwidth]{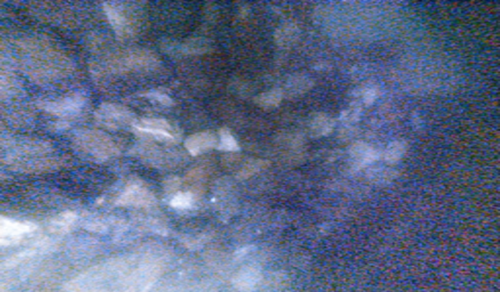}}
\hspace{0.1cm}
\subfigure{\includegraphics[width=0.09\textwidth]{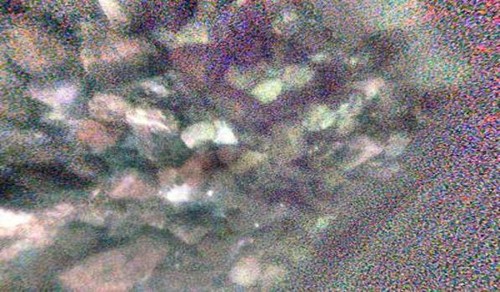}}
\vspace{-0.2cm}
\caption{\textbf{T}8-9: sunlight}
\label{fig:six_images_4}
\end{figure}
\vspace{-0.2cm}
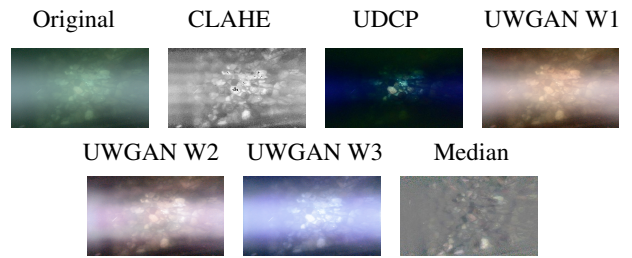
\begin{figure}[H]
\centering
\hspace{0.3cm}Original \hspace{0.8cm} CLAHE \hspace{1cm}UDCP \hspace{0.7cm} UWGAN W1 \hspace{0.5cm}
\subfigure{\includegraphics[width=0.09\textwidth]{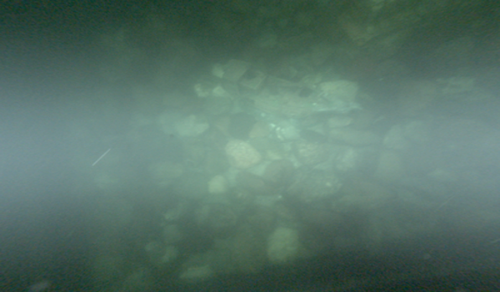}}
\hspace{0.1cm}
\subfigure{\includegraphics[width=0.09\textwidth]{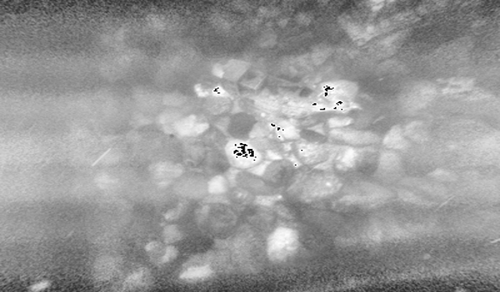}}
\hspace{0.1cm}
\subfigure{\includegraphics[width=0.09\textwidth]{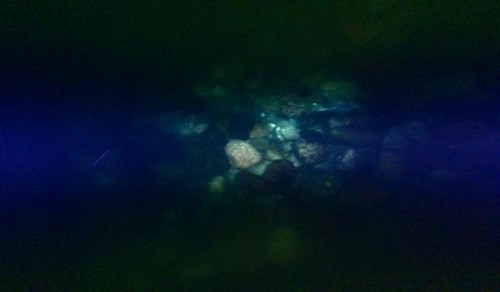}}
\hspace{0.1cm}
\subfigure{\includegraphics[width=0.09\textwidth]{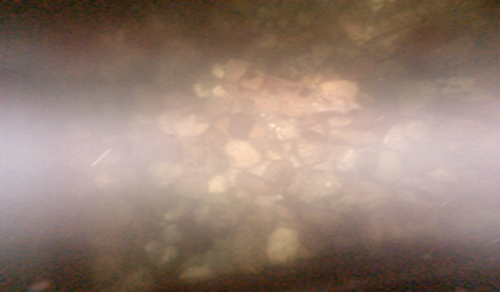}}
\hspace{5cm}
\begin{picture}(5,5)
\put(5,5){}
\end{picture}
\hspace{0.3cm} UWGAN W2 \hspace{0.2cm} UWGAN W3 \hspace{0.5cm} Median
\newline
\subfigure{\includegraphics[width=0.09\textwidth]{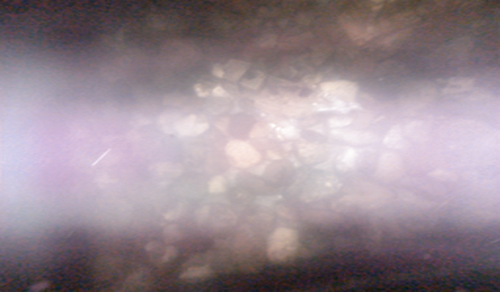}}
\hspace{0.1cm}
\subfigure{\includegraphics[width=0.09\textwidth]{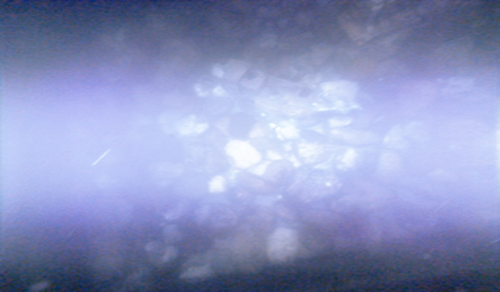}}
\hspace{0.1cm}
\subfigure{\includegraphics[width=0.09\textwidth]{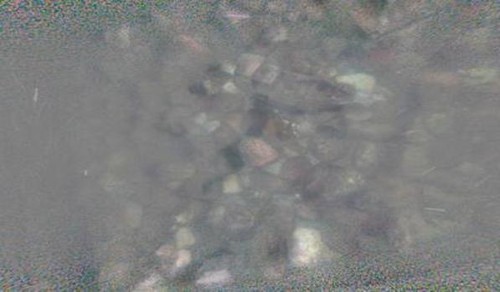}}
\vspace{-0.2cm}
\caption{\textbf{T}10-11: sunlight and artificial lights}
\label{fig:six_images_5}
\end{figure}
 \vspace{-0.1cm}
\begin{figure}[H]
\centering
\hspace{0.3cm}Original \hspace{0.8cm} CLAHE \hspace{1cm}UDCP \hspace{0.7cm} UWGAN W1 \hspace{0.5cm}
\subfigure{\includegraphics[width=0.09\textwidth]{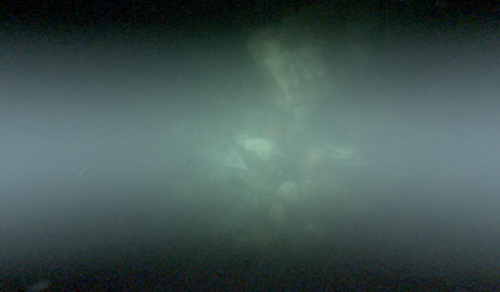}}
\hspace{0.1cm}
\subfigure{\includegraphics[width=0.09\textwidth]{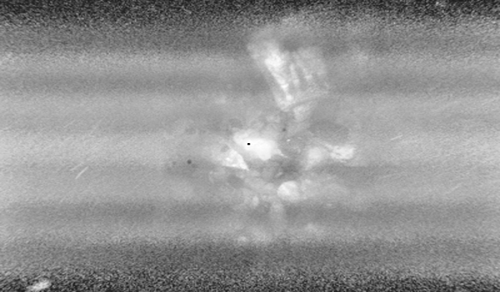}}
\hspace{0.1cm}
\subfigure{\includegraphics[width=0.09\textwidth]{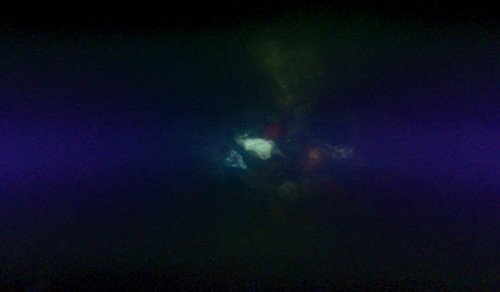}}
\hspace{0.1cm}
\subfigure{\includegraphics[width=0.09\textwidth]{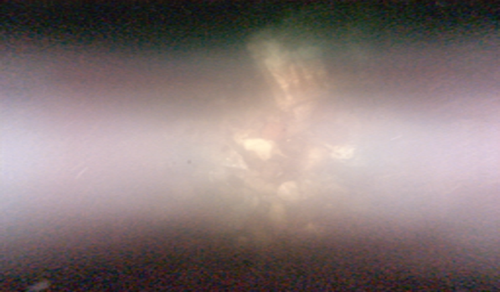}}
\hspace{5cm}
\begin{picture}(5,5)
\put(5,5){}
\end{picture}
\hspace{0.3cm} UWGAN W2 \hspace{0.2cm} UWGAN W3 \hspace{0.5cm} Median
\newline
\subfigure{\includegraphics[width=0.09\textwidth]{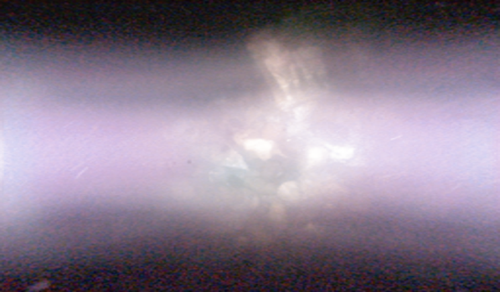}}
\hspace{0.1cm}
\subfigure{\includegraphics[width=0.09\textwidth]{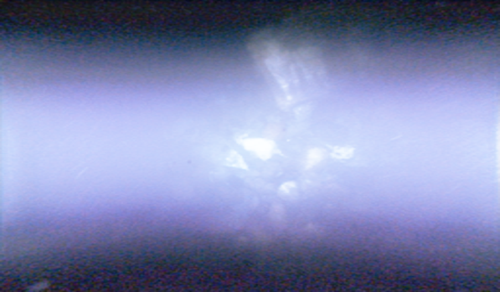}}
\hspace{0.1cm}
\subfigure{\includegraphics[width=0.09\textwidth]{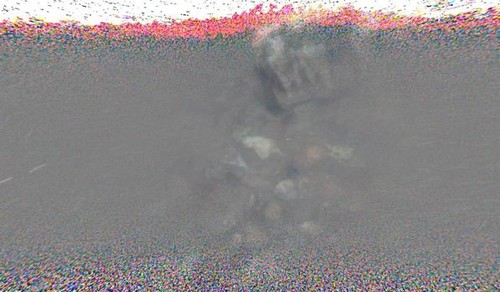}}
\caption{\textbf{T}12-13: artificial lights}
\label{fig:six_images_6}
\end{figure}
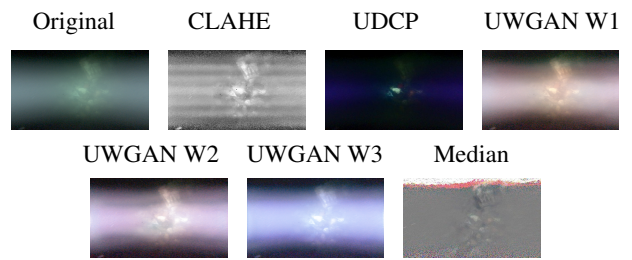

\end{document}